\documentclass{article}

\usepackage[preprint]{neurips_2026}
\usepackage{longtable}
\usepackage{tabularx}
\usepackage{multirow}
\usepackage{enumitem}
\usepackage{array}
\usepackage[most]{tcolorbox}
\usepackage{wrapfig}
\usepackage{listings}

\usepackage{xspace}

\usepackage[utf8]{inputenc} %
\usepackage[T1]{fontenc}    %
\usepackage{hyperref}       %
\usepackage{url}            %
\usepackage{booktabs}       %
\usepackage{amsfonts}       %
\usepackage{nicefrac}       %
\usepackage{microtype}      %
\usepackage{xcolor}         %
\usepackage{cleveref}
\usepackage[table]{xcolor}

\newcommand{\cOne}{\textsc{TaskPr}\xspace}
\newcommand{\cTwo}{\textsc{HumPr}\xspace}
\newcommand{\cThree}{\textsc{MemPr}\xspace}
\newcommand{\comp}{\textsc{Compactor}\xspace}

\definecolor{codebg}{HTML}{F7F7F7}
\definecolor{codeframe}{HTML}{CCCCCC}
\definecolor{kw}{HTML}{005CC5}      %
\definecolor{str}{HTML}{22863A}     %
\definecolor{com}{HTML}{6A737D}     %
\definecolor{num}{HTML}{B31D28}
\lstdefinestyle{echopy}{
  language=Python,
  basicstyle=\ttfamily\small,        %
  keywordstyle=\color{kw}\bfseries,
  stringstyle=\color{str},
  commentstyle=\color{com}\itshape,
  numberstyle=\tiny\color{com},
  showstringspaces=false,
  keepspaces=true,
  columns=fullflexible,
  breaklines=true,
  frame=single,
  rulecolor=\color{codeframe},
  backgroundcolor=\color{codebg},
  aboveskip=0pt,
  belowskip=0pt,
  xleftmargin=0.5em,
  xrightmargin=0.5em,
  tabsize=2,
}

\title{Simulating Human Memory with Language Models}

\author{%
  Qihan Wang\thanks{Primary contributors.} \\
  NYU \\
  \texttt{qw2488@nyu.edu} \\
  \And
  Nicholas Tomlin\footnotemark[1] \\
  NYU \\
  \texttt{n.tomlin@nyu.edu} \\
  \AND
  Michael Hu \\
  NYU \\
  \texttt{michael.hu@nyu.edu} \\
  \And
  Brian Dillon \\
  UMass Amherst \\
  \texttt{bwdillon@umass.edu} \\
  \And
  Tal Linzen \\
  NYU \\
  \texttt{linzen@nyu.edu} \\
}

\begin{document}

\maketitle

\begin{abstract}
\setcounter{footnote}{0}
Language models are increasingly being deployed as user simulators, but their memory is far more reliable than that of real users.
To measure this gap, we run a series of classic memory experiments from psychology on both humans and language models.
Across tasks, we find that out-of-the-box language models exhibit better memory than humans, even when prompted to imitate human behavior.
We then show that better prompting strategies and the use of a compactor can cause language models to forget content in a more human-like way. 
Using these methods, we show preliminary evidence that language models with human-like memory constraints can function as more effective user simulators in a downstream education task.
Finally, we release human reference data and benchmarks to support future work on simulating human memory with language models.\footnote{Code and data is available at \url{https://github.com/nickatomlin/simulating-memory}}
\end{abstract}

\section{Introduction}
\label{sec:intro}
Language models now match or surpass human performance on a wide range of benchmarks. As a result, models have become more similar to humans along certain dimensions, while becoming less similar along other dimensions. Compared to the language models of five, ten, or twenty years ago, today's models are remarkably capable of generating fluent sentences, maintaining semantic coherence over long texts, and performing commonsense reasoning; however, at the same time, today's models are often more helpful, more knowledgeable, and more verbose than typical humans.

While improving general-purpose capabilities may necessitate building models which diverge from typical human behavior---for most applications, for example, we may not want to replicate human reasoning biases \cite{eisape-etal-2024-systematic}---there are many reasons we might also want to build human-like language models. First and foremost for the context of the current work, more human-like models could serve as more useful user simulators, which could be used as a reward signal to train AI assistants which are better at collaborating with humans \citep{lintomlin2025usersim}. Second, human-like models can function as stand-ins for humans for the purpose of training people for certain jobs, e.g., the models can serve as mock students or patients \citep{pan2025tutorup, louie2026can}. Third, human-like language models can be used for population simulation, which can guide policy outcomes \citep{park2024generative}. Finally, building models that behave like humans has the opportunity to yield new insights into human cognition \citep{lake2017building,oh2025model}.

\begin{figure*}[t]
    \centering
    \includegraphics[width=\textwidth]{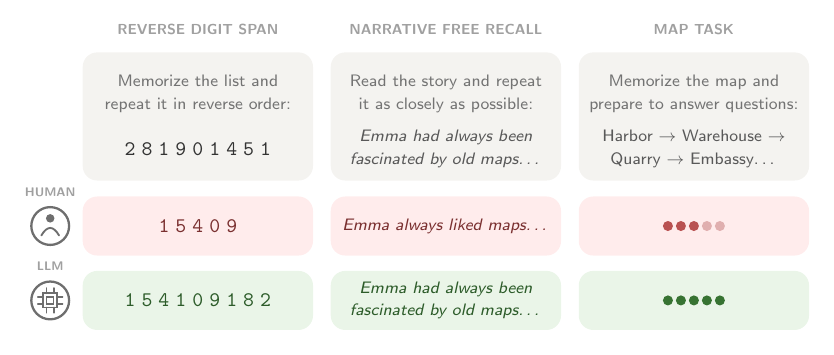}
    \caption{Three sample tasks from our benchmark. In reverse digit span, participants must read and recall lists of numbers in reverse order; in narrative free recall, participants must read short stories and repeat them as precisely as possible; in the map task, participants must memorize a map and answer questions about routes between locations. Humans are limited in their ability to remember the stimuli and complete each of these tasks, whereas frontier LLMs perform at ceiling.}
    \label{fig:teaser}
\end{figure*}

In this paper, we focus on \textit{memory} as one key dimension for building human-like language models. Recent improvements in long-context language modeling have yielded models which can effectively remember very long texts, because they can attend to any previous tokens in context \cite{armeni2022characterizing}. In contrast, decades of research on human memory have characterized systematic failures of recall, even over relatively short time horizons \cite{oberauer2018benchmarks}.  
Being able to simulate such human memory limitations would be useful if we want to, for instance, build models that can predict what a student would remember from a lesson, or what a programmer would remember from an interaction with a coding assistant.

In order to characterize the differences between humans and language models, we first develop a suite of tasks for evaluating memory (cf.~\Cref{fig:teaser}), grounded in tasks from the cognitive science and psychology literature. These tasks range from simple measures of working memory such as digit span, where subjects must remember a variable-length list of numbers, to more complex settings, such as one where subjects must memorize a map and answer questions about possible routes. We collect human data from $N{=}50$ participants and compare the distributions of human and language model scores under a variety of models and prompting strategies. We find that language models diverge from human behavior, achieving near-ceiling performance at all of the tasks, and they fail to closely simulate humans even when prompted to behave like humans with limited memory. There was no evidence that frontier models such as GPT-5.4 were better able to simulate human memory than less advanced models.

Next, we show that a simple approach rooted in well-established principles from human psychology can lead to better simulation of human memory. By first prompting a language model to summarize its context into four chunks \cite{cowan2001}, and then solve the tasks conditioned only on the content of these chunks, we find that language models achieve a closer fit to human scores on our suite of tasks. At the same time, this approach does not always lead to more humanlike \textit{patterns} of forgetting, pointing to ample space for improvement on this task.

Finally, we run a proof-of-concept experiment to show that models that can better simulate human memory may be more useful user simulators. Here, we design an education task where human subjects are presented with documents and asked comprehension questions. We modulate these documents by reading level, by the amount of redundancy, and by the amount of irrelevant information they contain, and attempt to predict which documents human subjects will best remember. We find that, when used as user simulators, out-of-the-box frontier models predict that students will be able to answer all questions, whereas the models with the most human-like memory can more effectively predict which documents will be useful to human learners. Even the most humanlike models we study, however, are far from perfect, again demonstrating substantial headroom on this task.

\section{Related Work}
\paragraph{Computational psycholinguistics.} 
LLMs' advantage in recalling past contexts is apparent at multiple time scales, and on multiple different types of input \citep{cao2025analyzing, oh2025model}. Compared to humans, LLMs demonstrate significantly enhanced verbatim memory of short idiomatic phrases \citep{rambelli2023frequent}, familiar longer-form documents \citep{mccoy2023much}, novel texts presented word-for-word \citep{vaidya2023humans}, and short patterned lists and sequences \citep{armeni2022characterizing}. 
However, models with more human-like memory can be useful both in modeling cognition, and in different applications \citep{wilcox2025bigger}. For example, constrained memory can improve language learning \citep{thamma2025human} and alignment with psycholinguistic measures of processing difficulty in humans \citep{de2024locally, clark2025linear}. Directly fine-tuning on experimental data can increase human-model alignment: \cite{binz2025foundation} introduced Centaur, a foundation model fine-tuned on data from cognitive psychology experiments. It successfully predicted the behavior of experimental participants, and had internal representations that were more aligned with neural data.  In a highly influential paper, Guti\'{e}rrez and colleagues showed that retrieval augmented generation systems inspired by the neurobiological hippocampal indexing theory can improve performance on multi-hop QA benchmarks \citep{gutierrez2024hipporag}. \cite{fountas2024human} similarly find performance improvements on long context benchmarks by organizing tokens in a model's context window into coherent `event' sequences, a feature of human episodic memory; for a review on episodic memory in LLMs, see \citep{dong2025towards}. Perhaps most relevantly for the present work, AI assistants that directly model human short-term and long-term memory interactions have been shown to improve question answering quality \citep{lee2024human,li2026himes}. 

\paragraph{Memory in foundation models.}

LLM memory includes implicit contextual memory, where information is retained within the context window or model activations \cite{dai2019transformer,bulatov2022recurrent}, and explicit long-term memory, where systems store, retrieve, and update information across interactions~\citep{wang2024memoryllm,wang2025m+}. 
Longer context windows do not guarantee reliable memory use, as models may fail to retrieve information \cite{liu2024lost}, motivating context-extension methods \cite{peng2023yarn,lu2025controlled} such as positional interpolation \cite{chen2023extending} and rotary embeddings \cite{su2024roformer}.
Beyond extending the context window, recent systems introduce model-updatable parameters \cite{wang2024self}, sparse embedding lookups \cite{cheng2026conditional,lin2025sparsememory}, and structured memory stores for persistent experience reuse~\citep{sun2025hierarchical,xu2025mem,rasmussen2025zep}. 
Most relevant to our work, \cite{de-langis-etal-2026-strong} evaluates LLMs on classic memory tasks such as digit span and N-back, but does not attempt to mimic human performance.
As memory and retrieval improve, evaluation has shifted toward long-horizon, multi-turn, and dynamic settings that test whether agents can manipulate information over extended interactions~\citep{maharana2024evaluating,wu2024longmemeval,hu2025evaluating,deshpande2025memtrack}.

\paragraph{User simulation with LLMs.}
Recent work uses LLMs as user simulators \cite{park2023generative,park2024generative}, leveraging their ability to instantiate personas and generate human-like actions to support large-scale behavioral experiments \cite{ge2024scaling}. 
One line of work studies how faithfully LLM personas reproduce human behavior \cite{venkit2026need,kang2025llm,wang2025user}. 
Another deploys LLM agents in social environments, from social-network simulations \cite{gao2023s3} to agent societies \cite{piao2025agentsociety,yang2024oasis}, modeling phenomena such as information diffusion \cite{liu2025mosaic} and negotiation \cite{zhu2025automated}.
Other work relies on user simulators for model evaluation \citep{10.1162/tacl_a_00679, barres2025tau} and as a reward signal for training \citep{sun2025training}, but has found the gap between simulated and real users to be a limiting factor \citep{zhou2026mind}.
Overall, LLMs' ability to precisely simulate human behavior remains uneven \citep{hu2025simbench,tjuatja-etal-2024-llms,lu2025can,wang2025llm}. This observation motivates our work, which aims to lay a foundation for more humanlike user simulators.

\section{The Human Memory Simulation Benchmark}
To assess whether language models can simulate human-like memory, we constructed a benchmark of ten memory tasks. We describe the tasks in our benchmark in \Cref{tab:tasks} and \S\ref{sec:tasks}. Then, in order to establish reference human performance, we collect human data for all ten tasks (\S\ref{sec:humandata}). Finally, we describe how we quantify LLMs' human-likeness on these tasks (\S \ref{sec:eval}).

\subsection{Tasks}
\label{sec:tasks}
Each task in our benchmark is designed to evaluate a different aspect of human memory. They include simple list recall tasks, such as the digit span task \citep{miller1956magical}, where participants are asked to remember lists of numbers (e.g., \texttt{1~8~7~9~1~2}), and the reverse digit span task \citep{gregoire1997effect}, where participants are expected to recall the same type of lists in reverse. Other tasks include narrative QA, which measures whether participants can remember temporal relationships between events in a story they have read, and the map task, which measures whether participants can memorize a map and describe paths to traverse it when the map is no longer presented on the screen. All tasks have text-only inputs and outputs, such that they can be performed by both humans and LLMs.

Whenever possible, our tasks are adapted from existing paradigms in psychology and cognitive science research on human memory. We modify some tasks to make them more suitable for LLMs; for example, in the classic digit span task numbers are presented auditorily, but we present them to humans visually, one at a time, in order to better match the language model version of the task. We provide short descriptions of all tasks in \Cref{tab:tasks}, and more detailed descriptions, along with the modifications we made to existing paradigms, in Appendix~\ref{app:task}.

\begin{table}
\small
\centering
\caption{List of tasks in our benchmark. We evaluate both humans and LLMs on each of these tasks and measure the distributional differences in their scores.}
\label{tab:tasks}
\begin{tabular}{p{13.3cm}}
\toprule

\textbf{Digit span.} Participants see sequences of digits and must recall them in the same order. Sequence length increases across trials, and performance is scored as the longest span that participants can fully reconstruct. This task is a classical measure of working memory capacity \cite{miller1956magical, gignac2015digit} and widely used in intelligence assessments \cite{silva2008development}, with typical human span around 7 $\pm$ 2 \cite{miller1956magical}.\vspace{0.2cm}\\

\textbf{Reverse digit span.} Participants see sequences of digits and must recall them in reverse order. Scoring is identical to digit span. Compared to digit span, this task imposes greater demands due to active manipulation of stored information. \cite{hilbert2014digit, gregoire1997effect} \vspace{0.2cm}\\

\textbf{N-back.} Participants see a sequence of letters and must identify whether each letter matches the one shown $n$ steps earlier (1-back, 2-back, 3-back). This is a widely used paradigm for studying cognitive control and executive function, including working memory processes. \cite{meule2017reporting, owen2005n} \vspace{0.2cm}\\

\textbf{Variable mapping.} Participants see a list of statements mapping names (variables) to locations (values). Each statement may either introduce a new variable (e.g., ``Alice lives in New York''), or may change the value of an existing variable (e.g., ``Alice moved to Boston''). After every two statements, participants must answer questions about who lives in which cities. This task relates to updating paradigms involving dynamic binding and replacement of information. \cite{schmiedek2009complex,ecker2010components} \vspace{0.2cm}\\

\textbf{Word recognition.} Participants are presented with a sequence of words, presented one at a time, and must judge whether each word has appeared before or not. The task ends after three errors and is scored by the number of correct responses. This task is based on continuous recognition paradigms. \cite{hockley1982retrieval} \vspace{0.2cm}\\

\textbf{Factual question answering (QA).} Participants read a Wikipedia passage and then answer multiple-choice questions without access to the passage. Performance is scored by accuracy. \vspace{0.2cm}\\

\textbf{Narrative QA.} Participants read a short story and then answer multiple-choice questions about event order without access to the passage. Performance is scored by accuracy. \vspace{0.2cm}\\

\textbf{Narrative free recall.} Participants read a short story and are then asked to reproduce it in as much detail as possible, without access to the passage. Performance is evaluated using embedding similarity. This task is adapted from naturalistic free recall paradigms and assesses memory of structured narratives. \cite{raccah2024naturalistic} \vspace{0.2cm}\\

\textbf{Map task.} Participants study a map and later answer route-based questions without access to the map. Performance is total accuracy across trials. This task relates to cognitive map formation and spatial relational memory. \cite{kitchin1994cognitive} \vspace{0.2cm}\\

\textbf{Craft task.} Participants study crafting recipes in a directed acyclic graph (\`a la Minecraft) and answer questions about how to create objects without access to the recipes. Performance is total accuracy across trials. This task evaluates the ability to draw inferences based on relational memory. \cite{ellenbogen2007human} \\
\bottomrule
\end{tabular}
\end{table}

\subsection{Human Data Collection}
\label{sec:humandata}
To collect human behavioral data, we developed an interactive web-based platform that implements all ten memory tasks. We adopted time constraints from earlier studies whenever possible (e.g., participants were given a fixed amount of time to read the text). We recruited $N{=}50$ participants via Prolific, restricting the sample to U.S.-based native English speakers with substantial platform experience (over 100 prior submissions), high approval rates (95–100\%), and no reported English language issues. Participants signed an informed consent form approved by our IRB. Each participant completed the tasks in a single session with a time limit of one hour, and was paid a flat rate of \$20. Forty out of fifty participants completed all ten tasks, and participants completed 9.56 out of 10 tasks on average. To mitigate order effects, the tasks were presented in a randomized order for each participant.

\subsection{Quantifying LLMs' Humanlikeness}
\label{sec:eval}
To quantify how closely LLMs simulate human memory, we use the Wasserstein distance, or earth mover's distance, between the distributions of human and LLM performance on the task; for example, in the digit span task, this would be a distribution over the maximal number of digits that the humans (as a population) were able to recall without error. Intuitively, Wasserstein distance captures how much ``effort'' is required to transform one distribution into the other, where effort depends on the probability mass that needs to be moved and the distance across which it needs to be moved.
Formally, let $X = \{x_i\}_{i=1}^n$ and $Y = \{y_j\}_{j=1}^m$ denote the human and LLM scores for a given task. The 1D Wasserstein distance is defined as:
\begin{equation*}
W_1(X, Y) = \int_{-\infty}^{\infty} \left| F_X(t) - F_Y(t) \right| dt,
\end{equation*}
which is simply the non-overlapping area between the empirical CDFs of humans and LLMs. Smaller $W_1$ indicates closer agreement between the human and LLM score distributions, with $W_1 = 0$ iff the distributions are identical.
To enable comparison across tasks, we normalize $X$ and $Y$ by the range of possible scores for each task. Finally, we convert $W_1$ into a similarity measure $\texttt{Humanlikeness} = 1 - W_1$, where larger values indicate closer human--model alignment.

\section{Methods}
We evaluate nine different LLMs, including models of varying sizes, reasoning and non-reasoning models, and a mixture of closed-weight and open-weight models. The full list of models is: GPT-5.4 \citep{singh2025openai}, Claude Opus 4.6 \citep{anthropic2026system}, Llama 3 8B Instruct, Llama 3.3 70B Instruct \citep{grattafiori2024llama}, Qwen3-8B (Standard), Qwen3-8B (Thinking), Qwen3-30B-A3B-Instruct, Qwen3-30B-A3B-Thinking, and Qwen3-Next-80B-A3B-Instruct \citep{yang2025qwen3}.

We present the tasks in Table \ref{tab:tasks} to these models in four ways. The first three only vary the prompt (for the exact prompts, see Appendix~\ref{app:methods}). The Task Prompt (\cOne) describes the task without making reference to simulating humans; this prompt elicits the model's out-of-the-box behavior on each task. The Human Prompt (\cTwo) explicitly instructs the model to simulate a human participant in a psychology experiment. It then describes the conditions under which humans take the test---\textit{e.g.,} for reading comprehension tasks, we tell the model that humans have five minutes to read the passages. Finally, the Memory Prompt (\cThree) is similar to \cTwo, but also reminds the model that humans have limited memory and therefore sometimes make mistakes. %

 In addition to these three prompt-only conditions, we construct an agent \cite{yao2023react,sumers2024cognitive} that we refer to as \comp{}. This agent explicitly implements a human-like working memory bottleneck via a key-value memory module. 
Humans are typically understood to have a working memory capacity of about four chunks, each of which can consist of a compressed, complex
concept \cite{cowan2001}. We mirror this constraint architecturally using the memory module.

\begin{wrapfigure}{r}{0.48\textwidth}
    \centering
    \vspace{-\intextsep}
    \vspace{-0.25cm}
    \begin{lstlisting}[style=echopy, caption={\comp pseudocode.}, label={lst:reflexion}]
def encode(LM, inp):
    memory = {}
    while not LM.finished(memory):
        LM.call({
            write_memory,
            delete_key
        })
    return memory

def compactor(LM, inp, query):
    memory = encode(LM, inp)
    answer = LM.recall(memory, query)
    return answer
\end{lstlisting}
\vspace{-0.3cm}
\end{wrapfigure}

\paragraph{Memory module.} The agent is given a key-value store with
a hard capacity of $K=4$ entries. Keys are short string labels chosen
by the model (e.g.\ \texttt{"characters"}, \texttt{"theme"}) and values
are free-form abstractive summaries. 
The memory store is exposed to the model as two tool-use
functions in the OpenAI tool-calling format:
\texttt{write\_memory(key, value)}, which inserts or overwrites an
entry (rejected if the store is full and \texttt{key} is new), and
\texttt{delete\_key(key)}, which removes an entry.

\comp answers questions using information stored in the memory module only.
In the \emph{encode} phase, the model sees the full information a human would see and performs tool calls to add information to memory. In the \emph{recall} phase, the model is re-invoked with a recall prompt that contains only the key-value contents and the question. 
We ablate \comp in Appendix~\ref{app:ablation}, comparing it to an unstructured
summarizer agent that receives the same input but
produces a single free-form prose summary instead of $K$ keyed chunks.
Both conditions answer from the summary only, isolating the contribution of the capacity-limited working memory tool from the more general effect of summarizing the input before answering.

\begin{figure}[t]
    \centering
    \includegraphics[width=\linewidth]{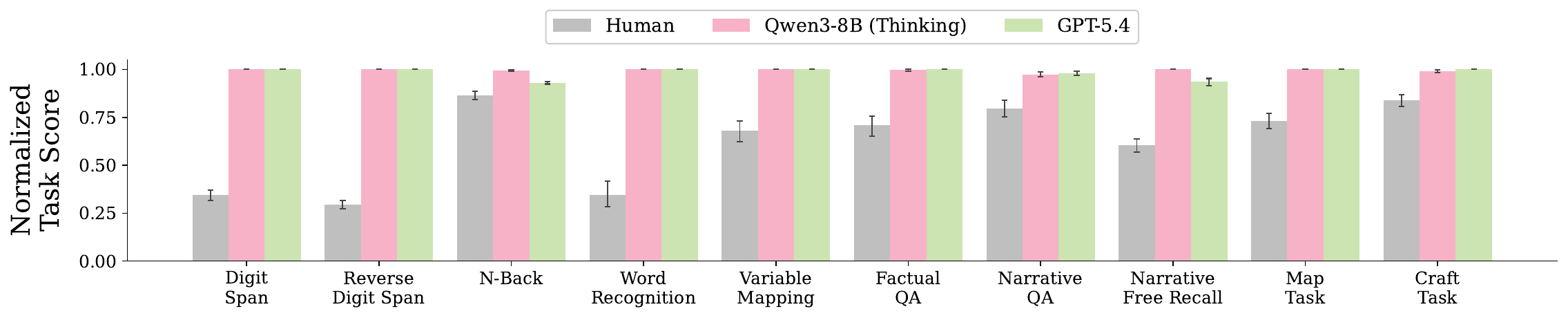}
    \caption{Human--model performance comparisons across tasks and for two representative models, Qwen3-8B and GPT-5.4 (results from other models are equally close to ceiling performance). The plot shows the models' performance on the tasks using the baseline prompt \cOne. Scores are normalized such that 1.0 represents perfect accuracy on each task. Error bars indicate bootstrapped 95\% confidence intervals.}
    \label{fig:similarity-grid}
\end{figure}

\section{Results}

\paragraph{Default LLM behavior deviates substantially from human memory.}
In the baseline condition \cOne, which does not instruct the model to simulate humans, all models diverge dramatically from human memory: in all ten tasks, models with the \cOne prompt exhibit near-perfect memory, (\Cref{fig:similarity-grid}). For example, in the digit span task, all models perfectly remember digits spans up to length twenty, far surpassing the human performance average of 6.88 digits. This pattern holds regardless of model size or whether the model uses chain-of-thought reasoning.

\paragraph{Explicit human-simulation prompting has limited and inconsistent impact.}
Introducing explicit instructions to simulate human participants (\cTwo) or reminding models of human memory limitations (\cThree) generally does not substantially improve human-model similarity. Across most tasks, $\Delta$\cTwo and $\Delta$\cThree---the change in performance in \cTwo and \cThree compared to \cOne---are near-zero or even negative for the majority of models. To use the example of the digit span task again, $\Delta$\cTwo is exactly $0$ for all models, though we do see improvements from $\Delta$\cThree for a small subset of models (e.g., $+0.55$ for Qwen3-30B-Thinking). In the variable mapping task, the human simulation prompts actually \emph{reduce} similarity (e.g., $\Delta$\cTwo = $-0.13$ and $\Delta$\cThree = $-0.13$ for GPT-5.4).
Narrative free recall is the only setting where the human simulation prompt leads to consistent and substantial improvements across many models; most models show large $\Delta$\cThree (e.g., $+0.16$ for GPT-5.4, $+0.32$ for Qwen3-8B-Standard, +0.3 for Qwen3-Next-80B). Overall, though, when we consider the full range of tasks and models, we find that prompting alone is insufficient to reliably induce human-like memory behavior.

\paragraph{Inclusion of an explicit memory module yields more human-like results.}
In contrast to prompt-only interventions, \comp, a language
agent that explicitly enforces a human-like working memory bottleneck via a key-value memory module, produces substantially more human-like score distributions across models and tasks (Figure~\ref{fig:similarity-grid}). Improvement is especially pronounced for the working-memory tasks: digit span, reverse digit span, N-back, and word recognition. In the digit span task, where all models exhibit identical non-human-like behavior with the baseline prompt \cOne, \comp yields large improvements across models, with $\Delta$\comp reaching $+0.61$ for Qwen3-8B Standard. 
Word recognition shows the same pattern, with consistent gains across models (e.g., up to $\Delta\comp = +0.26$), in stark contrast to the near-zero or inconsistent effects of the human simulation prompts. For tasks that depend on longer-term memory, such as factual QA and narrative QA, the effect is more variable across models. In factual QA, $\Delta\comp =0$ for Claude Opus 4.6, but reaches $+0.29$ for Llama 3 8B. Overall, we conclude that \comp, which explicitly implements cognitive principles of human memory, is more effective than prompting models to behave like humans with limited memory.

\begin{figure}[t]
    \centering
    \includegraphics[width=\linewidth]{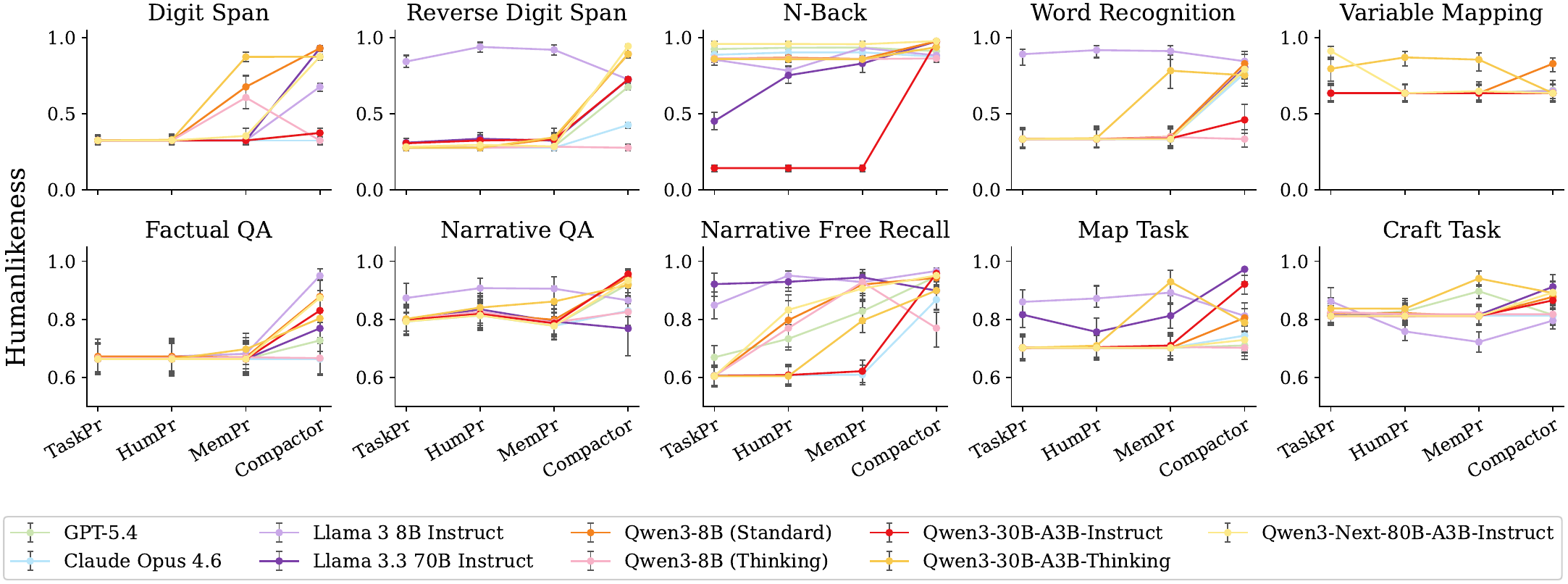}
    \caption{Human--model similarity across models and prompt conditions. Error bars indicate bootstrapped 95\% confidence intervals.}
    \label{fig:similarity-grid}
\end{figure}

\section{Analysis}

In the previous section, we showed that simple prompting strategies (\cOne, \cTwo, and \cThree) are insufficient to reliably induce human-like memory behavior across tasks. We next investigate if in-context examples of human behavior help language models better simulate human memory (\S \ref{sec:icl-transfer}), and if language models' detailed forgetting patterns are similar to those of humans (\S \ref{sec:forgetting-analysis}).

\subsection{Examples of Limited Working Memory Do Not Transfer Across Tasks}
\label{sec:icl-transfer}

Can we improve the models' humanlikeness by providing examples of humans' performance on the task? We evaluate GPT-5.4 on two representative tasks: digit span and word recognition. In addition to the three base prompting conditions (\cOne, \cTwo, and \cThree), we introduce two new few-shot conditions, which we call \textit{in-domain} and \textit{out-of-domain}. For both few-shot settings, we randomly sample five human participants from our collected dataset and include their entire trial responses as demonstrations in the prompt. These examples are prefaced with the instruction: \textit{``Here are example results from previous human participants.''} This exposes the model to realistic human behavior, including errors and termination patterns. 
In the \textit{in-domain} condition, demonstrations come from the same task (e.g., digit span examples for digit span). In the \textit{out-of-domain} condition, the demonstrations come from the other task (e.g., word recognition examples for digit span, and vice versa). In the out-of-domain case, we clarify in the prompt that these examples correspond to a different task. Both tasks probe working memory, while \textit{out-of-domain} tests whether human-like behavior can transfer across tasks that test the same underlying cognitive function.

\begin{figure}[t]
    \centering
    \includegraphics[width=\linewidth]{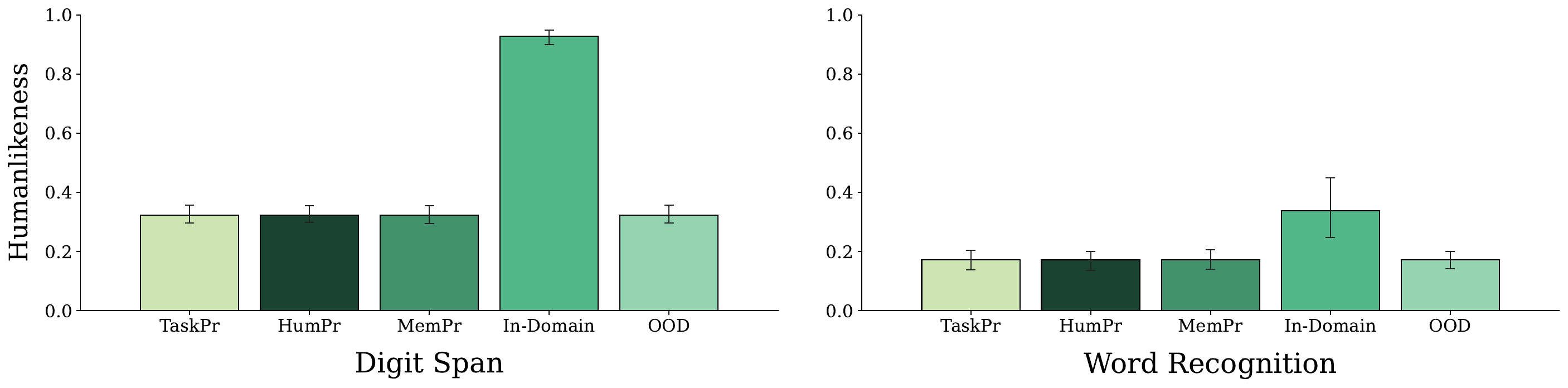}
    \caption{Few-shot prompting experiments. Human--model similarity across prompting conditions for GPT-5.4 on two working memory tasks. In-domain few-shot examples substantially improve alignment with human behavior, while out-of-domain examples yield little to no improvement over the baseline. Error bars denote 95\% bootstrapped confidence intervals.}
    \label{fig:analysis}
\end{figure}

For both tasks, we observe that adding same-task few-shot examples consistently improves humanlikeness relative to all other prompting conditions, as shown in Figure \ref{fig:analysis}.
In contrast, adding examples from a different task does not yield comparable improvements. When the model is given human transcripts from a different task that also evaluate working memory, performance remains similar to the \cOne baseline. This suggests that the benefit of few-shot prompting is not attributable to exposure to generic human-like mistakes, but rather to elements of task-specific structure. 

These results highlight a key limitation of in-context learning (ICL) for human simulation: improvement from few-shot learning is limited to in-domain settings, a phenomenon also observed in prior work \cite{min-etal-2022-rethinking,liu-etal-2022-makes,mueller-etal-2024-context}. 
More broadly, since ICL does not generalize even in this narrow benchmark setting, it is unlikely that ICL is a robust tool for simulating human memory in the wild.

\subsection{Language Models and Humans Forget Different Things}
\label{sec:forgetting-analysis}
The scoring criteria for our task measure the limits of perfect recall, as is standard in the human memory literature: if a participant or a model recalled one of ten digits incorrectly, the trial is marked as incorrect. In an analysis of the detailed responses, we find that even when LLMs do make mistakes on the same trials that humans do, those mistakes are often not the same types of mistakes. In digit span, for example, \cThree and \comp both increase the coarser humanlikeness score, which captures the maximal number of digits that the models can recall perfectly, but they exhibit specific forgetting patterns that differ from those of humans. Averaging across models in the \cThree condition, we find that 87\% of incorrect predicted spans match the true span length (compared to only 55\% for humans). Models in the \cThree condition frequently reproduce the entire sequence with one or two incorrect digits in the middle, e.g., predicting \mbox{\texttt{6 7 2 6 8 8 2 7 4 9 9}} instead of \mbox{\texttt{6 7 2 6 0 8 2 7 4 9 9}}; this pattern is unusual in humans. In contrast, the \comp frequently predicts truncated versions of the original sequence, e.g., predicting \mbox{\texttt{3 1 0 3 4}} instead of \mbox{\texttt{3 1 0 3 4 1 3 1}}---again an unusual pattern for humans. These differences are reflected in the conditional rate of errors, i.e., the probability that a model will predict an incorrect digit given that it has predicted the previous digit incorrectly. We find that $p(\text{err} \mid \text{prev wrong}) = 71.2\%$ for humans, $56.1\%$ for \cThree, and $94.1\%$ for \comp. These results suggest that more work is needed to build models with human-like \textit{patterns} of forgetting.

\section{Proof-of-Concept Application: Reranking Educational Documents}
\label{sec:application}
Finally, we consider an application where modeling human performance is highly relevant: teaching. Training AI assistants using LLM user simulators that misrepresent human cognitive capabilities---for example, ones that retain information much more readily than humans do---is likely to result in AI assistants that overestimate human capabilities and are therefore ill-suited for teaching humans. As a proof of concept assessing whether human memory simulation could lead to models that are more useful for teaching, we design a controlled reading comprehension task with multiple versions of each document that vary in difficulty and memory demands, and evaluate whether different user simulators can correctly identify the version of the document that humans can best remember. 

\paragraph{Materials.}
We generate ten fictional biographies (cf. Appendix~\ref{app:application_data_construction} for details), each paired with ten multiple-choice questions. We construct four variants of each document, all of which have the same answer-relevant content: a version written at a middle-school reading level; a version at a higher reading level, with more advanced vocabulary and syntax; a redundant version, where facts required to answer each question appear twice, making key information easier to remember; and a distractor version, which introduces irrelevant information that is not needed to answer the questions, increasing memory load and, we hypothesize, making the document harder relative to the baseline biography. Across all variants, the question set and answer key are held fixed.

\paragraph{Human and LLM evaluation.}
We collect data from 100 human participants recruited on Prolific, each of whom reads one of the versions of one of the documents, and answers questions about the text after the text is no longer visible. We also evaluate LLMs on the same documents and questions, in the four conditions described above (\cOne, \cTwo, \cThree, and \comp); for additional details on human data collection, see Appendix~\ref{app:humanapplication}. 

We compute humanlikeness as before (\Cref{sec:eval}). To evaluate the potential usefulness of the user simulators in a hypothetical teaching application where it is beneficial to select the version of a text that is most likely to be understood by a student, we also compute \emph{pairwise reranking accuracy}, as follows. We construct pairwise trials by sampling two documents from the set of $10 \times 4$ biographies without replacement.
For each document, we compute the mean accuracy based on human or model responses. The preferred document is defined as the one with higher mean accuracy, with ties broken randomly. We then measure alignment by comparing the model’s preference with the humans' preference; the pairwise reranking accuracy is defined as the fraction of trials where the two agree. All results are estimated over 10,000 sampled trials.

\begin{figure}[t]
    \centering
    \includegraphics[width=\linewidth]{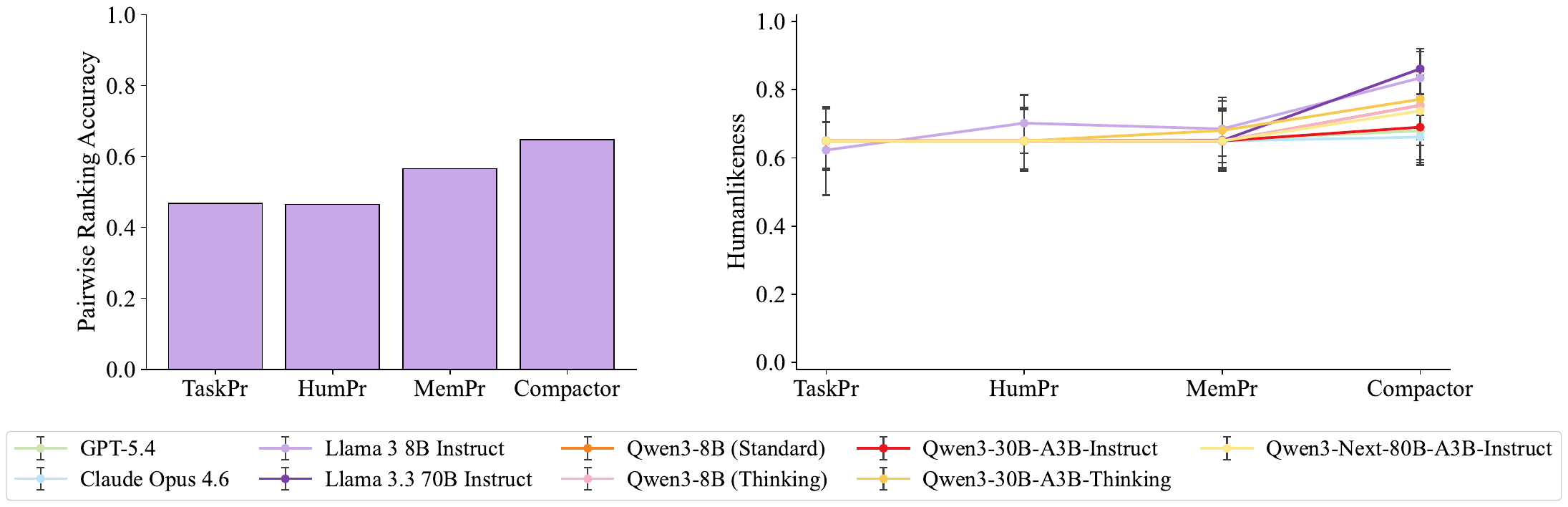}
    \caption{Proof-of-concept application: rereanking educational documents. \textbf{Left}: Pairwise reranking accuracy for the best-performing model (Llama 3 8B Instruct). This accuracy measures the proportion of pairs of documents for which the user simulator and the human made a greater number of errors on the same document. An accuracy of 50\% corresponds to random selection.
\textbf{Right}: Humanlikeness of LLM user simulators across models and prompt conditions.}
    \label{fig:application}
\end{figure}

\paragraph{Results.} As in the ten benchmark tasks, \comp shows higher humanlikeness compared to the three prompt-only conditions, although the gains are not uniform across all models (\Cref{fig:application}). For example, Llama 3 8B Instruct, the best-performing model on the closely related factual QA task from the benchmark, achieves $\Delta\comp = +0.21$. This suggests that the explicit humanlike memory mechanism implemented in \comp is helpful in estimating how difficult humans would find a document. 
We also find preliminary evidence that \comp can help determine the relative difficulty of two documents for a human: for example, Llama 3 8B Instruct achieves a $+0.180$ absolute gain in accuracy compared to \cOne, and $+0.147$ gain compared to random selection. However, these improvements are not consistent across models, and even the best-performing model leaves substantial room for further gains.

\section{Conclusion}

User simulators can serve as an important tool for scaling up the evaluation and training of AI models that interact with humans. But for these simulators to be useful, they must be realistic: to use a simulator to evaluate a model's ability to teach humans, for example, the simulator needs to be limited in its ability to absorb, process and retain information in the same way that humans are \cite{lintomlin2025usersim}.

The present study focuses on a particular cognitive capability: memory. Typical LLMs' memory is far superior to that of humans---they can repeat far longer lists of numbers than a human could, for example \cite{armeni2022characterizing,oh2025model}. As such, they are unlikely to be useful user simulators out of the box. We confirm this discrepancy in a comparison between LLMs and human participants in a battery of ten memory tasks that we develop. This discrepancy generally persists even when the LLMs are explicitly prompted to simulate a human with limited memory capacity. %
Alignment with human memory improves when we implement an explicit memory limitation, where the LLM must compact (summarize) the context using only four slots, the approximate number of memory slots that humans have \cite{cowan2001}. Finally, we propose an evaluation setting for user simulators in a teaching context, where the number of errors made by the simulator in comprehension questions is used to decide how to present the information to users; we show that simulators with more humanlike memory perform better in this task. 

The prompting and compaction approaches we implement, while generally improving over the baseline, are only the first step toward simulators with realistic humanlike memory. All of these approaches resulted in detailed error patterns that differed from those of humans, and in the experiment where models were expected to compare texts by their memorability, the accuracy of our best model was less than 70\%. This leaves substantial room for improving the humanlikeness of user simulators on our benchmark; one promising approach may be to make the context compaction mechanism more similar to human memory in its details. At a higher level, we advocate for user simulators that have the same resource limitations as humans, including in other domains (for example, reasoning), and for cognitive benchmarks that can measure alignment between simulators and humans.

\section*{Acknowledgments}

We thank the NYU Computation and Psycholinguistics Lab for comments and suggestions. TL is supported by the National Science Foundation (NSF) under Cooperative Agreement No. 2433429, “NSF AI Research Institute on Interaction for AI Assistants (ARIA)”, and by the National Institute of Biomedical Imaging and Bioengineering under grant R01EB038873. MYH is supported by the NSF Graduate Research Fellowship. BD is supported by the Samuel F. Conti fellowship from the University of Massachusetts, Amherst and by NSF IIS-2504954 to the University of Massachusetts, Amherst. This work was supported in part through the NYU IT High Performance Computing resources, services, and staff expertise.

\bibliographystyle{unsrt}
\bibliography{references}

\appendix

\section{Limitations}
\label{app:limitations}
In this section, we describe additional limitations of our work, which are avenues for future work.

\paragraph{Coverage of memory types.} While our suite of tasks is intended to provide a diverse coverage of memory types, it does not capture all aspects of human memory; for example, semantic memory and visual memory are not tested in our benchmark. This limitation is largely an intentional choice, as we focused on tasks with ``text in, text out" structure so that it would be easier to make comparisons between humans and language models. Future work may study additional types of human memory.
\paragraph{Modeling distributional behavior.} Our proposed method, \comp, is meant to be a first step toward simulating human memory and is not intended as a complete solution to this problem. Although we evaluate models based on their distributional similarity to human responses, our approach is not explicitly designed to capture distributional differences in human responses. Future work could take advantage of techniques like verbalized sampling \citep{zhang2025verbalized} or persona-conditioned prompting \citep{park2023generative} to obtain more diverse sets of model responses, which might yield a better fit to human data.

\section{Task Information}
\subsection{Full Task Descriptions}
\label{app:task}
\paragraph{Digit span.}
The digit span task assesses short-term verbal working memory by presenting participants with sequences of digits one at a time and requiring them to reproduce each sequence in the same order after presentation ends. Across trials, the digit strings gradually increase in length, placing greater demands on working memory. In this task, participants receive two sequences at each length, and the task stops after two errors at the same length. Performance is scored by best span, defined as the longest sequence length at which the participant correctly reproduces both sequences. This is a classical and widely used measure of working memory capacity in cognitive psychology \cite{miller1956magical, gignac2015digit} and is widely used in standardized intelligence assessments (e.g., WAIS) \cite{silva2008development}. Previous research shows that humans have a digit span of 7 $\pm$ 2 \cite{miller1956magical}.

\paragraph{Reverse digit span.}
The reverse digit span task assesses working memory by presenting participants with sequences of digits one at a time and requiring them to reproduce each sequence in reverse order after presentation ends. Across trials, the digit strings gradually increase in length, increasing the demand on working memory. In this task, participants receive two sequences at each length, and the task stops after two errors at the same length. Performance is scored by best span, defined as the longest sequence length at which the participant correctly reproduces both sequences. Compared to the digit span, the reverse version places greater demands on working memory as it requires active manipulation of the stored information, and is widely used as a measure of executive working memory capacity in cognitive psychology and neuropsychology \cite{hilbert2014digit, gregoire1997effect}. 

\paragraph{N-back.}
The N-back task assesses working memory by presenting a continuous sequence of letters one at a time, and requiring participants to decide, on each eligible trial, whether the current letter matches the one shown a fixed number of steps earlier. In this implementation, participants perform three scored blocks at increasing levels of demand: 1-back, 2-back, and 3-back. Successful performance requires continuously updating recently presented information. Performance is scored using accuracy percentage, defined as the proportion of correct responses out of all eligible responses. The N-back task is one of the most popular experimental paradigms of working memory \cite{meule2017reporting}. It requires on-line monitoring, updating, and manipulation of remembered information, placing substantial demands on multiple core processes of working memory \cite{owen2005n}.

\paragraph{Variable mapping.}
The variable mapping task assesses working memory by requiring participants to maintain and update associations between individuals and locations across a sequence of statements. Participants read brief sentences one at a time indicating where each person lives, with some trials introducing changes in location, and after every two statements they are asked to identify the city where a named person currently lives. Successful performance depends on tracking the most recent person-location pairing while replacing outdated information when updates occur. The task ends when participants make a mistake. Because the task is relatively short and prone to mistakes, each participant is allowed for three attempts, and only the highest score is retained as the final score. The score is defined as the largest number of active person-location associations correctly maintained at the time of a correct response, with the final score equal to the best such value achieved across the three attempts. This task is closely related to working memory updating paradigms, which require maintaining and dynamically updating arbitrary bindings while discarding outdated information \cite{schmiedek2009complex,ecker2010components}. 

\paragraph{Word recognition.}
The word recognition task presents words one at a time and requires participants to judge whether each word has appeared earlier in the list or is being shown for the first time. The task ends after three incorrect responses. Performance is scored by the number of correct responses achieved before the task ends. This task is closely related to continuous recognition paradigms, in which stimuli are presented in a continuous stream and participants must identify previously encountered items online \cite{hockley1982retrieval}. Such tasks rely on the ability to maintain recently presented items in an active state and compare incoming stimuli against them, engaging short-term working memory processes.

\paragraph{Factual QA.}
The factual QA task assesses reading comprehension and factual memory by giving participants three minutes to read a passage derived from Wikipedia and then requiring them to answer ten multiple-choice questions after the text is removed. In this implementation, the reading material is based on Wikipedia source documents, and GPT-5.1 is used to produce a passage of about 600 words for presentation in the task. Performance is scored as the accuracy of the questions. This task can be viewed as a form of episodic memory assessment, in which participants encode the information in the reading and later retrieve it after the stimulus is no longer available.

\paragraph{Narrative QA.}
The narrative QA task assesses memory for story content by giving participants three minutes to read a 600 words narrative and then requiring them to answer ten multiple-choice questions after the text is removed. The narratives are generated materials paired with question sets varying difficulties by LLMs. Questions are about the order of the events happened in the narrative. Performance is scored as the accuracy of the questions. This task can be viewed as a form of episodic memory assessment where participants need to encode a few events \cite{cohn2022narratives}.

\paragraph{Narrative free recall.}
The narrative free recall task assesses memory for extended verbal material by giving participants five minutes to read a story and then requiring them to recall as much of the story as possible after the text is hidden. Successful performance depends on retaining narrative details, wording, and overall structure well enough. In this implementation, performance is summarized by the similarity between the given material and recalled text, evaluated by a BLEU score and an embedding-based similarity score computed using \texttt{sentence-transformers/all-MiniLM-L6-v2}. The narrative free recall task is adapted from the naturalistic free recall paradigm \cite{raccah2024naturalistic}, in which participants encode extended narrative material and subsequently reconstruct it from memory after the stimulus is removed. This task probes episodic memory for temporally structured events, requiring participants to retain and retrieve narrative details, ordering, and high-level semantic structure.

\paragraph{Map task.}
The map task assesses memory by requiring participants to study a map of locations and the roads connecting them, forming a non-directional graph, and then answer route-based questions after the map is removed. In this task, participants complete three trials, each consisting of one minute to memorize the map. After the map disappears, participants answer five multiple-choice questions about how to travel between locations using the available road connections. Performance is scored as the total accuracy across all three trials, out of fifteen questions in total. This task is closely related to cognitive map formation \cite{kitchin1994cognitive}, where individuals encode relational spatial structure and later use it to support navigation.

\paragraph{Craft task.}
The craft task assesses memory by requiring participants to study a set of materials and crafting rules that form a directed acyclic graph (DAG), and then answer questions about how items combine after the rules are removed. In this implementation, participants complete three trials, each consisting of one minute of study followed by five memory-based questions about the crafting system. Performance is scored as the total accuracy across all three trials, out of fifteen questions in total. The craft task evaluates relational memory by requiring participants to encode a structured system of dependencies between items and later use this representation to answer questions \cite{ellenbogen2007human}.

\subsection{Dataset Construction and Generation}
\label{app:datasetgeneration}
\paragraph{Factual QA.} To construct the dataset, we start from Wikipedia source documents and filter them for suitability. Specifically, we prompt GPT-5.1 to assess whether a document is appropriate for a coherent $\sim$600-word reading passage (e.g., avoiding pages dominated by lists, tables, or highly fragmented structure). For selected documents, we then use the LLM to rewrite the content into a paragraphized passage of approximately 600 words and to generate ten corresponding multiple-choice questions. The resulting passage--question pairs constitute the factual QA dataset used in our experiments.

\paragraph{Narrative QA.}
To construct the narrative QA dataset, we first prompt GPT-5.1 to generate coherent narrative stories under controlled constraints, including a target length (approximately 600 words), a fixed number of events, and a requirement of clear temporal progression within a single continuous storyline. We then prompt the LLM to generate ten multiple-choice questions for each story. The questions are designed to emphasize temporal reasoning (e.g., event order, before/after relations) while also including a subset of factual event questions. All questions are answerable solely based on the story, with exactly four options and a single correct answer. We further control for difficulty and answer distribution by encouraging medium-to-hard questions and approximately balanced correct options across A--D. The resulting story--question pairs constitute the narrative QA dataset used in our experiments.

\paragraph{Narrative free recall.}
For the narrative free recall task, we directly use the “Naturalistic Free Recall” dataset \cite{raccah2024naturalistic}. This dataset consists of 4 naturalistic narrative stimuli paired with high-fidelity human recall transcripts collected from hundreds of participants. In the original study, participants listened to spoken narratives and were then asked to recall the stories in as much detail as possible after the stimulus was removed. In our implementation, we adapt the task to a text-based setting: participants are given five minutes to read the story and are then asked to type as much of the story as they can remember after the text is removed.

\paragraph{Map task.}
To construct the map task, we use GPT-5.1 to generate synthetic navigation environments and corresponding questions. Each instance consists of a set of locations connected by edges, forming a graph that participants must memorize. We control task difficulty by varying the number of locations, with three levels corresponding to 4, 5, and 6 locations. For each map, the LLM generates a set of connectivity relations between locations, followed by multiple-choice questions that require reasoning over valid paths between a start and a goal location. Each question presents candidate routes, of which only one corresponds to a valid path in the underlying graph.

\paragraph{Craft task.}
To construct the craft task, we use GPT-5.1 to generate synthetic crafting systems and corresponding questions. Each instance defines a set of items and a collection of crafting rules, where pairs of items can be combined to produce new items, forming a directed acyclic graph over item dependencies. We control task difficulty by varying the number of items, with three levels corresponding to 5, 6, and 7 items. For each task, the LLM generates natural language descriptions of crafting rules, followed by multiple-choice questions that require reasoning over valid compositions and dependencies among items. Each question presents candidate answers, of which only one is consistent with the underlying crafting rules.

\section{Human Experiment Details}
\label{app:human_data}

\paragraph{Experimental platform.}

\begin{figure}[t]
    \centering
    \includegraphics[width=1\linewidth]{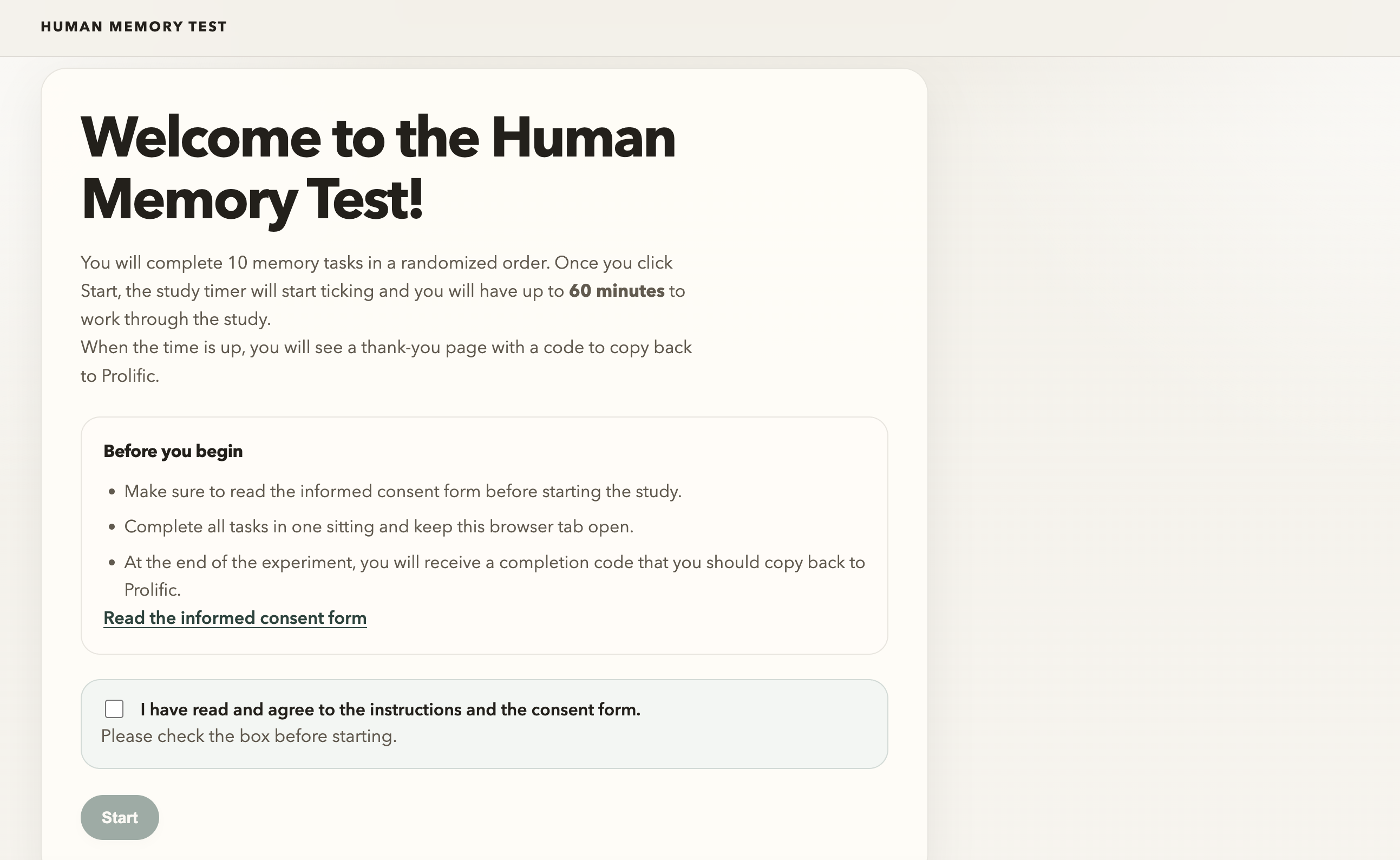}
    \includegraphics[width=1\linewidth]{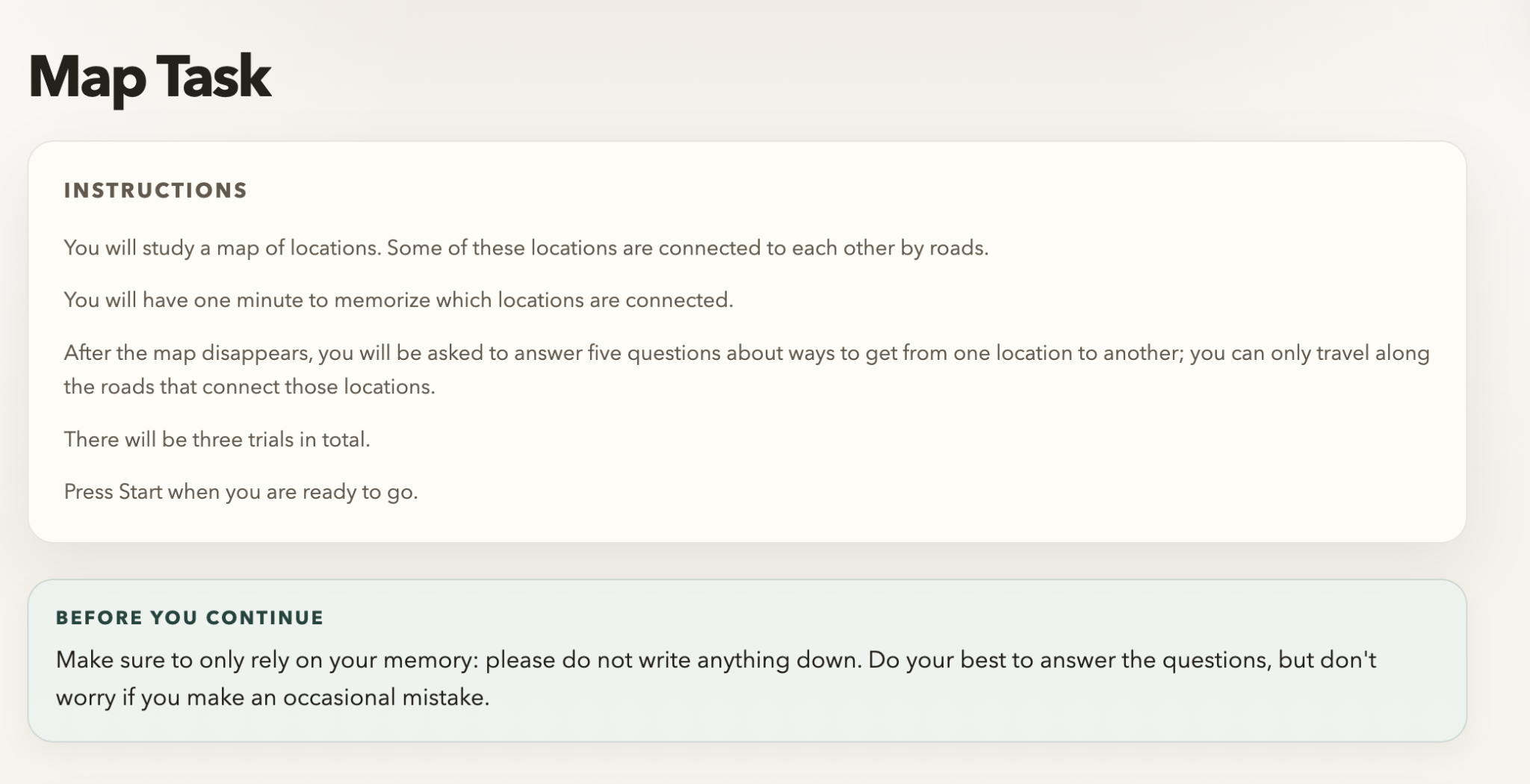}
    \caption{The main instruction and example user interfaces from our experimental platform. Top: Main instruction on the welcome page. Bottom: Map task, where participants memorize spatial connectivity and answer route queries.}
    \label{fig:ui_examples}
\end{figure}

To collect human behavioral data, we developed an interactive web-based platform that implements all ten memory tasks. The frontend of the platform was deployed using Heroku. All task events and responses were logged in real time and stored in Amazon S3. Each task was presented as a self-contained page with clear instructions, controlled timing (e.g., fixed reading durations), and input constraints. Figure~\ref{fig:ui_examples} shows example main instruction and task instruction displays.

\paragraph{Recruitment.}
We recruited 50 participants through Prolific using stringent pre-screening criteria. Participants were restricted to U.S.-based native English speakers with substantial platform experience (100--100{,}000 prior submissions), high approval rates (95--100\%), and no self-reported language-related disorders. Each participant was compensated \$20.

\paragraph{IRB approval.}
\label{app:irb}
All participants provided informed consent prior to participation. The study protocol was reviewed and approved by our Institutional Review Board. We do not anticipate any significant risks to participants.

\subsection{Task Instructions}
In this section, we present the instructions shown to human participants for each task. All instructions were displayed on a web-based interface, and participants initiated each task by clicking a \textit{Start} button.

\paragraph{Digit span.}
You will see a sequence of digits presented one at a time. Your task is to remember the digits in the exact order they appear. After the sequence ends, type the digits in the same order and press Submit. Type the digits without spaces (e.g., 3917). The sequences will gradually become longer. Try to remember them as accurately as possible.

\paragraph{Reverse digit span.}
You will see a sequence of digits presented one at a time. Your task is to remember the digits and enter them in reverse order. After the sequence ends, type the digits from last to first and press Submit. Type the digits without spaces. For example, if the sequence is 3917, you should enter 7193. The sequences will gradually become longer. Try to remember them as accurately as possible.

\paragraph{N-back.}
In this task, a letter is \textit{Same} if it matches the letter immediately before it, and \textit{Different} otherwise. The first letter does not require a response. After that, respond to each letter as Same or Different. The next letter will not appear until you respond. You will first complete a short practice block with feedback, followed by the scored block.

\paragraph{Word recognition.}
Words will appear one at a time. For each word, decide whether it has already appeared earlier in the list. Select \textit{Old} if the word has appeared before, and \textit{New} otherwise. The first word is always new. You have three strikes, and the task ends after three incorrect answers.

\paragraph{Variable mapping.}
You will see a series of sentences describing where people live. Try to remember where each person lives. Note that people may move to a new city. After every two sentences, you will be asked a question of the form: ``Where does [Name] live?'' Respond with the city where the person currently lives.

\paragraph{Factual QA.}
You will have three minutes to read a passage, after which the text will disappear. You will then answer ten questions about the text.

\paragraph{Narrative QA.}
You will have three minutes to read a passage, after which the text will disappear. You will then answer ten questions about the text.

\paragraph{Narrative free recall.}
You will first have five minutes to read a story. After the story is hidden, type as much as you remember in a text box and submit your response. Try to recall the story as precisely as possible, using the original wording when possible.

\paragraph{Map task.}
You will study a map of locations connected by roads. You will have one minute to memorize which locations are connected. After the map disappears, you will answer five questions about how to travel between locations using the available roads. There are three trials in total.

\paragraph{Craft task.}
You will study a set of materials and crafting rules. You will have one minute to memorize how items combine. After the rules disappear, answer five questions from memory. There are three trials in total.

\section{LLM Experiments}
\subsection{Benchmark}
\label{app:benchmark}
The Human Memory Simulation Benchmark is a reproducible benchmarking suite designed to evaluate the extent to which language models can simulate human memory behavior.
The benchmark and all associated resources are available at \url{https://github.com/nickatomlin/simulating-memory}.
The benchmark comprises ten core tasks described in Appendix~\ref{app:task}, including tasks evaluating short-term memory (e.g., forward and reverse digit span, n-back), episodic memory (e.g. factual QA and narrative QA), and relational memory (e.g., craft task and map task). The benchmark is designed with reproducibility and flexibility in mind. All tasks are implemented within a unified command-line framework and operate over versioned stimulus data, with random seeds recorded to ensure deterministic replay of experiments. Configuration files specify model parameters, task settings, and output structure, enabling consistent evaluation across models and conditions.

\subsection{Compute Resources}
\label{app:compute_resource}
All experiments are conducted via API-based access to pretrained language models (OpenAI API and OpenRouter API). As such, no local model training is performed. The primary computational cost comes from API inference calls.

\subsection{Methods}
\label{app:methods}
\subsubsection{Prompting-Based Approaches}
To evaluate how closely language models simulate human memory, we evaluate each model under three prompting conditions, denoted as \cOne, \cTwo, and \cThree.  These conditions differ in how explicitly the model is instructed to behave like a human participant. For each model and condition, we run 50 independent trials for each task to align with the scale of the human experiment. 

\paragraph{\cOne: General description.}
The model is given only a task description without any reference to humans. This prompt is designed to elicit the model’s default behavior on the task. We refer to this prompt as the \textit{LLM prompt}.

\paragraph{\cTwo: Explicit human simulation.}
In this condition, the model is explicitly instructed to simulate a human participant. The prompt consists of a prefix instruction followed by a task description written from a human perspective:
\begin{quote}
\textit{“You are simulating a human participant in a psychology experiment.”}
\end{quote}
This is followed by a \textit{human prompt}, which is obtained by transcribing the original participant instructions into a form suitable for model input. The human prompt closely mirrors the instructions shown to human participants, including task descriptions, constraints, and examples.

\paragraph{\cThree: Explicit simulation with limited memory reminder.}
In this condition, we further encourage human-like behavior by explicitly reminding the model of human memory limitations. The prompt consists of:
\begin{quote}
\textit{“You are simulating a human participant in a psychology experiment. Behave as much like a realistic human as possible. Remember that humans have limited memory and therefore sometimes make mistakes.”}
\end{quote}
This is followed by the same \textit{human prompt} used in \cTwo.

Notably, \cTwo can be viewed as a strict subset of \cThree.

\paragraph{Prompt construction.}
Across all tasks, we construct two types of prompts: (1) the \textit{LLM prompt}, which contains only the task description and is used in \cOne, and (2) the \textit{human prompt}, which mirrors the instructions shown to human participants and is used in \cTwo and \cThree. This design allows us to isolate the effect of human-simulation instructions and memory constraints on model behavior.

\subsubsection{\comp}
Our \comp is an LLM agent with the ability to interact with a key-value memory store. The agent is given the following prompt, detailing how to store contents in the memory store.

\paragraph{Prompt.} You are simulating a human participant in a psychology experiment on working memory.
  You have a key-value memory store with at most 4 slots, reflecting the ~4-chunk
  limit of human short-term memory (Cowan, 2001).
  Use \texttt{write\_memory} and \texttt{delete\_key} to maintain the key-value store while doing the original task.
  Each slot should hold ONE chunk — a small bundle of information a person would bind together
  because it feels meaningfully connected (a name with its role, a group of related items or
  numbers, one gist). When the task asks for verbatim retrieval of a sequence, a human will
  form meaningful chunks of 1–3 items, starting from the beginning. NEVER pack a long run of items into one slot. Once your slots are filled, accept
  that the rest will be lost. Compress realistically, and behave as a real human would:
  imperfect and sensitive to what seems important.

\paragraph{Inference.} At inference time, the \comp model is prompted with ``Your working memory currently contains: \{wm\_contents\}'' and no longer has access to the original document or task stimuli.

\subsection{Prompts}

We provide the prompts used in all tasks. For each task, we define two components: (1) the \textit{LLM prompt}, which describes the task without reference to humans, and (2) the \textit{human prompt}, which mirrors the instructions given to human participants.

\paragraph{Digit span.}
\begin{itemize}[leftmargin=*, itemsep=4pt]
    \item \textbf{LLM prompt:} You will see a sequence of digits presented one at a time. Your task is to remember the digits in the exact order they appear. After the sequence ends, type the digits in the same order. The sequences will gradually become longer. Try to remember them as accurately as possible.
    
    \item \textbf{Human prompt:} The human will see a sequence of digits presented one at a time. Their task is to remember the digits in the exact order they appear. Then, the sequence will disappear. After the sequence disappears, they will be asked to type the digits in the same order as they appeared. The sequences will gradually become longer. They will be asked to remember them as accurately as possible.
\end{itemize}

\paragraph{Reverse digit span.}
\begin{itemize}[leftmargin=*, itemsep=4pt]
    \item \textbf{LLM prompt:} You will see a sequence of digits presented one at a time. Your task is to remember the digits and enter them in reverse order. After the sequence ends, type the digits from last to first. The sequences will gradually become longer. Try to remember them as accurately as possible.  

    For example, if the digits are the following: [4, 8, 2]  
    You should answer: press <<2>>. press <<8>>. press <<4>>.
    
    \item \textbf{Human prompt:} The human will see a sequence of digits presented one at a time. Their task is to remember the digits and enter them in reverse order. Then, the sequence will disappear. After the sequence disappears, they will be asked to type the digits from last to first. The sequences will gradually become longer. They will be asked to remember them as accurately as possible.  

    For example, if the digits are the following: [4, 8, 2]  
    You should answer: press <<2>>. press <<8>>. press <<4>>.
\end{itemize}

\paragraph{N-back.}
\begin{itemize}[leftmargin=*, itemsep=4pt]
    \item \textbf{LLM prompt:} You will be shown a sequence of letters. After every letter, you will decide whether it matches the letter one turn back. In each block, respond with "no response" to the first letter. Once enough letters have appeared, respond to each new letter as "same" or "different".  

    Example: A → A → B → C → C  
    Responses: no response, same, different, different, same
    
    \item \textbf{Human prompt:} The human will be shown a sequence of letters. After every letter, the human will decide whether it matches the letter one turn back. In each block, the human is asked to respond with "no response" to the first letter. Once enough letters have appeared, respond to each new letter as "same" or "different".  

    Example: A → A → B → C → C  
    Responses: no response, same, different, different, same
\end{itemize}

\paragraph{Variable mapping.}
\begin{itemize}[leftmargin=*, itemsep=4pt]
    \item \textbf{LLM prompt:} You will see a series of sentences describing where people live. Try to remember where each person lives. Pay attention: people will occasionally move to a new city. After every two sentences, you will be asked: “Where does [Name] live?” Respond with the city where the person currently lives.
    
    \item \textbf{Human prompt:} The human will see a series of sentences describing where people live. Sentences are presented one at a time. Each sentence disappears before the next sentence or question appears, and previous sentences are not visible. The human is asked to remember where each person lives. Pay attention: people will occasionally move to a new city. After every two sentences, the human will be asked: “Where does [Name] live?” Respond with the city where the person currently lives.
\end{itemize}

\paragraph{Word recognition.}
\begin{itemize}[leftmargin=*, itemsep=4pt]
    \item \textbf{LLM prompt:} Words will appear one at a time. For each word, decide whether it has already appeared earlier in the list. Respond with "old" if the word has appeared before and "new" otherwise.
    
    \item \textbf{Human prompt:} The human will see words one at a time. Each word disappears before the next word appears. For each word, the human must decide whether it has already appeared earlier in the list. They respond with "old" if the word has appeared before and "new" otherwise.
\end{itemize}

\paragraph{Factual QA.}
\begin{itemize}[leftmargin=*, itemsep=4pt]
    \item \textbf{LLM prompt:} Read a passage, and then answer ten questions.
    
    \item \textbf{Human prompt:} The human will have three minutes to read a passage, after which the text will disappear. The human will then be asked to answer ten questions about the text.
\end{itemize}

\paragraph{Narrative QA.}
\begin{itemize}[leftmargin=*, itemsep=4pt]
    \item \textbf{LLM prompt:} Read a passage, and then answer ten questions.
    
    \item \textbf{Human prompt:} The human will have three minutes to read a passage, after which the text will disappear. The human will then be asked to answer ten questions about the text.
\end{itemize}

\paragraph{Narrative free recall.}
\begin{itemize}[leftmargin=*, itemsep=4pt]
    \item \textbf{LLM prompt:} Read a story, then recall the story as precisely as possible using the same words when possible. For example, if the story is in first person, you should also use first person.
    
    \item \textbf{Human prompt:} The human will have five minutes to read a story. The story will then be hidden. The human will be asked to type as much as they remember. They are asked to recall the story as precisely as possible using the same words when possible. For example, if the story is in first person, they should also use first person.
\end{itemize}

\paragraph{Map task.}
\begin{itemize}[leftmargin=*, itemsep=4pt]
    \item \textbf{LLM prompt:} You will study a map of locations and roads. Some locations are connected by roads. Memorize which locations are connected, then answer five questions about possible routes. There are three trials in total.
    
    \item \textbf{Human prompt:} The human will study a map of locations. Some locations are connected by roads. The human will have one minute to memorize which locations are connected, after which the map will disappear. The human will then answer five questions about how to travel between locations using only the available roads. There are three trials in total.
\end{itemize}

\paragraph{Craft task.}
\begin{itemize}[leftmargin=*, itemsep=4pt]
    \item \textbf{LLM prompt:} You will study a set of materials and crafting rules. Memorize how items combine, then answer five questions from memory. There are three trials in total.
    
    \item \textbf{Human prompt:} The human will study a set of materials and crafting rules. The human will have one minute to memorize how items combine, after which the rules will disappear. The human will then answer five questions from memory. There are three trials in total.
\end{itemize}

\section{Ablation Study for \comp}
\label{app:ablation}
\begin{figure}[t]
    \centering
    \includegraphics[width=0.6\linewidth]{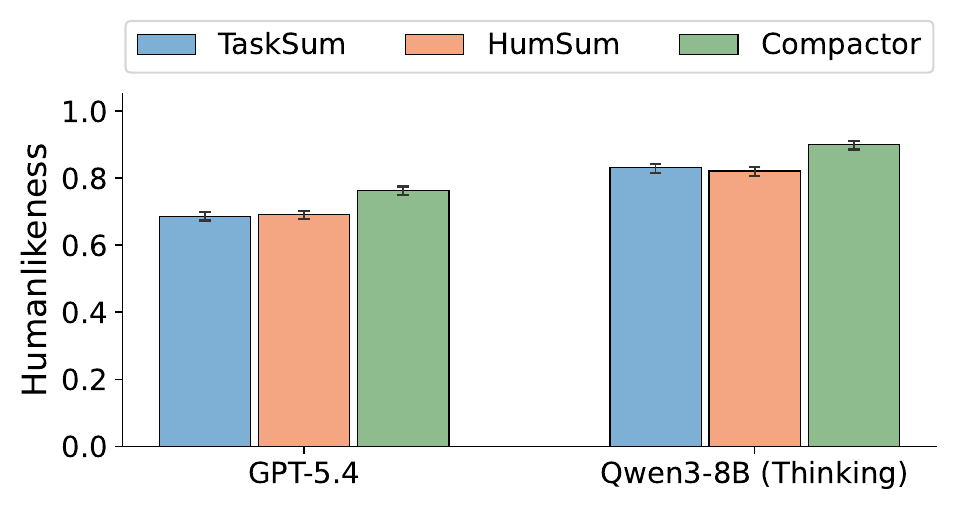}
    \caption{Ablation results for the \comp. We find that \comp, which uses a key-value memory store, achieves higher humanlikeness than a simpler method which prompts a model to produce an abstractive summary of its context before completing the task.}
    \label{fig:ablation}
\end{figure}

We run an ablation to compare our \comp model, which writes memories to a key-value store, to a simpler model which directly produces an abstractive summary of its context. We design two summarization methods: (1) \textsc{TaskSum}, which produces an abstractive summary and then is directly prompted to complete the task, and (2) \textsc{HumSum}, which produces an abstractive summary and then is prompted to simulate human-like behavior on the task. We compare \comp with both summarizer methods on two representative models, Qwen3-8B (Thinking) and GPT-5.4, finding that \comp achieves higher humanlikeness when averaging across tasks, as shown in \Cref{fig:ablation}. 

\paragraph{Summarizer prompt:} You will first be shown material to remember. Produce a concise abstractive summary of it — keep the summary short (prefer brief, dense prose; aim for roughly a paragraph, not a transcript). You will later have to answer questions using ONLY your summary, so make sure the summary captures what you'll need for the task above.

\section{Additional Details: Reranking Educational Documents}

This appendix provides additional details on the experiment described in \Cref{sec:application}.

\label{app:application}
\subsection{Materials}
\label{app:application_data_construction}
We construct the dataset for this experiment using a synthetic data pipeline based on prompt templates. We first generate base documents and question sets, and then derive controlled variants from the same underlying content. 
We generate documents based on a predefined list of professions. For each sampled profession, we call GPT-5.1 to produce one fictional biography together with exactly ten multiple-choice questions. The generation follows a fixed prompt template, ensuring consistent structure and topic across all documents.

From each base biography, we construct three additional variants using separate rewrite prompts, while keeping the questions and answer keys unchanged. The reading-level condition rewrites the document using more complex vocabulary and phrasing, increasing linguistic difficulty without changing the underlying facts. The redundant condition repeats or rephrases answer-relevant information so that key facts appear multiple times, making them easier to retrieve. The distractor condition introduces additional irrelevant but contextually plausible details, increasing memory load through interference while preserving the correct answers.

We release all data and generation code to facilitate future research.

\subsection{Human Experiment}
\label{app:humanapplication}

The human experiment largely follows the setup of the main benchmark, with minor modifications to accommodate the reading comprehension task. We recruited 100 participants through Prolific using stringent pre-screening criteria. Each participant is assigned a single document under one reading condition. Participants are given three minutes to read the document, after which the text is removed, and they answer ten multiple-choice questions based on memory. Participants are required to complete the entire task within eight minutes. Compensation is set to \$2.70 per task, corresponding to an approximate hourly rate of \$20.

\subsection{LLM Experiment}
\begin{figure}[t]
    \centering
    \includegraphics[width=0.6\linewidth]{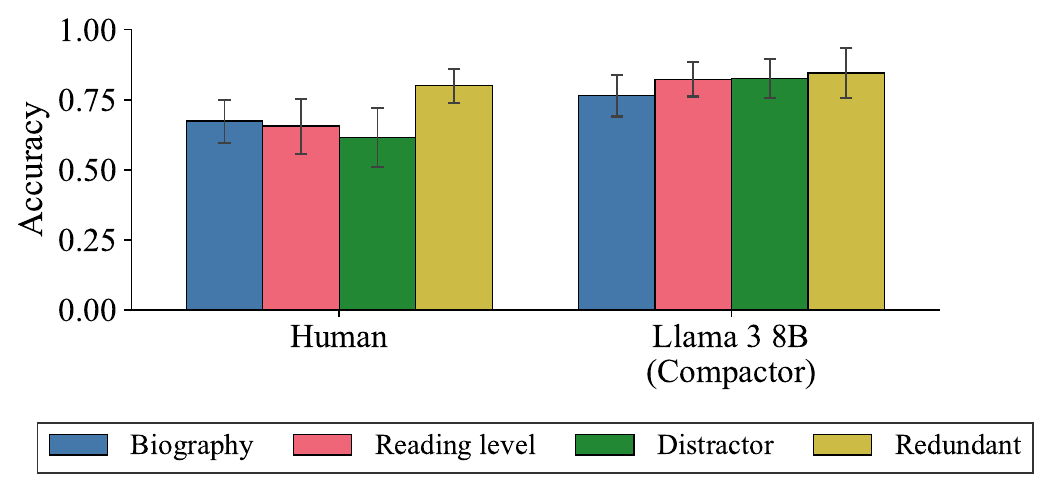}
    \caption{Although Llama 3 8B with \comp achieves higher pairwise accuracy across conditions, the relative ordering of document difficulty still differs from human behavior, suggesting there remains room for improvement.}
    \label{fig:application_breakdown}
\end{figure}

The evaluation protocol follows the same setup as the factual QA task described in the main benchmark. All prompting strategies and formats are identical to those used in the main experiments, including \cOne, \cTwo, \cThree and \comp. For each model and condition, we run 100 independent trials for each task to align with the scale of the human experiment. We report the average QA accuracy for each of the four document types in \Cref{fig:application_breakdown}.

\section{Additional Results}
We present additional results in \Cref{tab:performance-raw-bold} and \Cref{tab:wasserstein-norm-delta}.

\begin{table*}[t]
\centering
\small
\caption{Performance comparison. \cOne is the baseline. \cTwo, \cThree, and \comp are the alternative conditions. Statistically significant differences from \cOne are shown in \textbf{bold}.}
\label{tab:performance-raw-bold}
\vspace{0.1cm}
\setlength{\tabcolsep}{3pt}
\resizebox{\textwidth}{!}{%
\begin{tabular}{llrrrrrrrrrr}
\toprule
Task & Cond. & Human & Claude Opus 4.6 & Llama 3 8B & Llama 3.3 70B & GPT-5.4 & Qwen3 8B NT & Qwen3 8B T & Qwen3 30B Inst & Qwen3 30B Think & Qwen3 Next 80B \\
\midrule
\multirow{4}{*}{Digit Span}
 & \cOne & 6.88 & 20.00 & 20.00 & 20.00 & 20.00 & 20.00 & 20.00 & 20.00 & 20.00 & 20.00 \\
 & \cTwo & \cellcolor{gray!15} & 20.00 & 20.00 & 20.00 & 20.00 & 20.00 & 20.00 & 20.00 & 19.84 & 20.00 \\
 & \cThree & \cellcolor{gray!15} & 20.00 & 20.00 & 20.00 & 20.00 & \textbf{13.30**} & 12.60 & 20.00 & \textbf{4.34**} & 19.38 \\
 & \comp & \cellcolor{gray!15} & 20.00 & \textbf{0.00*} & \textbf{7.00*} & \textbf{19.00*} & \textbf{8.00*} & \textbf{7.00*} & \textbf{19.00*} & \textbf{9.00*} & \textbf{9.00*} \\
\midrule
\multirow{4}{*}{Reverse Digit Span}
 & \cOne & 5.90 & 20.00 & 9.08 & 19.36 & 20.00 & 20.00 & 20.00 & 19.46 & 20.00 & 19.86 \\
 & \cTwo & \cellcolor{gray!15} & 20.00 & \textbf{7.12**} & 18.92 & 20.00 & 20.00 & 20.00 & 19.06 & 20.00 & 19.66 \\
 & \cThree & \cellcolor{gray!15} & 20.00 & \textbf{7.50**} & 19.18 & 19.88 & \textbf{18.86**} & 20.00 & 19.02 & \textbf{18.64*} & 19.78 \\
 & \comp & \cellcolor{gray!15} & \textbf{17.00*} & \textbf{0.00*} & \textbf{0.00*} & \textbf{12.00*} & \textbf{9.00*} & \textbf{8.00*} & \textbf{11.00*} & \textbf{8.00*} & \textbf{4.00*} \\
\midrule
\multirow{4}{*}{N-Back}
 & \cOne & 0.866 & 0.976 & 0.717 & 0.315 & 0.931 & 0.996 & 1.000 & 0.000 & 1.000 & 0.904 \\
 & \cTwo & \cellcolor{gray!15} & \textbf{0.960*} & \textbf{0.642**} & \textbf{0.618**} & 0.927 & 0.995 & 0.995 & 0.000 & 1.000 & 0.901 \\
 & \cThree & \cellcolor{gray!15} & \textbf{0.959*} & \textbf{0.795**} & \textbf{0.696**} & 0.923 & 0.999 & 0.995 & 0.000 & 1.000 & 0.901 \\
 & \comp & \cellcolor{gray!15} & \textbf{0.988*} & 0.747 & \textbf{0.874***} & \textbf{0.951**} & \textbf{0.833***} & \textbf{0.890***} & \textbf{0.848***} & \textbf{0.926***} & \textbf{0.868**} \\
\midrule
\multirow{4}{*}{Word Recognition}
 & \cOne & 34.49 & 100.0 & 42.06 & 100.0 & 100.0 & 100.0 & 100.0 & 100.0 & 100.0 & 100.0 \\
 & \cTwo & \cellcolor{gray!15} & 100.00 & 35.96 & 100.00 & 100.00 & 100.00 & 100.00 & 100.00 & 99.46 & 100.00 \\
 & \cThree & \cellcolor{gray!15} & 100.00 & 32.58 & 100.00 & 100.00 & 98.66 & 100.00 & 99.48 & \textbf{55.80**} & 100.00 \\
 & \comp & \cellcolor{gray!15} & \textbf{40.20***} & \textbf{18.94***} & \textbf{30.74***} & \textbf{19.20***} & \textbf{44.76***} & \textbf{46.56***} & \textbf{87.52**} & \textbf{9.78***} & \textbf{33.38***} \\
\midrule
\multirow{4}{*}{Variable Mapping}
 & \cOne & 6.79 & 10.00 & 10.00 & 10.00 & 8.84 & 10.00 & 10.00 & 10.00 & 4.56 & 6.56 \\
 & \cTwo & \cellcolor{gray!15} & 10.00 & 10.00 & 10.00 & \textbf{9.96**} & 10.00 & 10.00 & 10.00 & \textbf{7.12**} & \textbf{10.00**} \\
 & \cThree & \cellcolor{gray!15} & 10.00 & 10.00 & 10.00 & \textbf{10.00**} & 10.00 & 10.00 & 10.00 & \textbf{7.88**} & \textbf{9.88**} \\
 & \comp & \cellcolor{gray!15} & 10.00 & 9.88 & 10.00 & 8.68 & \textbf{8.56***} & \textbf{6.88***} & 10.00 & \textbf{10.00***} & \textbf{10.00***} \\
\midrule
\multirow{4}{*}{Map Task}
 & \cOne & 10.98 & 15.00 & 12.88 & 13.68 & 15.00 & 15.00 & 15.00 & 15.00 & 15.00 & 15.00 \\
 & \cTwo & \cellcolor{gray!15} & 15.00 & 12.66 & \textbf{14.44**} & 15.00 & 15.00 & 15.00 & 14.98 & 14.90 & 15.00 \\
 & \cThree & \cellcolor{gray!15} & 15.00 & \textbf{12.36**} & 13.78 & 15.00 & 15.00 & 15.00 & 14.92 & \textbf{11.98**} & 15.00 \\
 & \comp & \cellcolor{gray!15} & \textbf{14.44*} & \textbf{8.50***} & \textbf{11.12***} & 14.92 & \textbf{13.78***} & \textbf{13.78***} & \textbf{11.68***} & \textbf{14.00***} & \textbf{14.70**} \\
\midrule
\multirow{4}{*}{Craft Task}
 & \cOne & 12.57 & 15.00 & 10.58 & 14.98 & 15.00 & 14.90 & 14.80 & 15.00 & 14.74 & 15.00 \\
 & \cTwo & \cellcolor{gray!15} & 15.00 & \textbf{8.92**} & 15.00 & \textbf{14.84*} & 14.90 & 15.00 & 15.00 & 14.74 & 15.00 \\
 & \cThree & \cellcolor{gray!15} & 15.00 & \textbf{8.48**} & 14.96 & \textbf{13.00**} & 15.00 & 15.00 & 15.00 & \textbf{13.12**} & 15.00 \\
 & \comp & \cellcolor{gray!15} & 15.00 & \textbf{9.62**} & \textbf{13.80***} & 14.96 & \textbf{14.06***} & \textbf{14.18***} & \textbf{14.46***} & \textbf{14.14***} & \textbf{14.04***} \\
\midrule
\multirow{4}{*}{Narrative QA}
 & \cOne & 7.96 & 9.80 & 9.22 & 9.78 & 9.80 & 9.78 & 9.80 & 9.76 & 9.74 & 9.80 \\
 & \cTwo & \cellcolor{gray!15} & 9.66 & 8.94 & \textbf{9.52*} & 9.66 & 9.62 & 9.40 & 9.62 & 9.48 & 9.66 \\
 & \cThree & \cellcolor{gray!15} & 9.90 & 8.94 & 9.80 & 9.90 & 9.76 & 9.60 & 9.84 & \textbf{9.26**} & 9.90 \\
 & \comp & \cellcolor{gray!15} & \textbf{9.52*} & \textbf{6.70***} & \textbf{5.72***} & \textbf{8.58***} & \textbf{7.36***} & \textbf{7.60***} & \textbf{7.74***} & \textbf{7.26***} & \textbf{7.48***} \\
\midrule
\multirow{4}{*}{Factual QA}
 & \cOne & 7.08 & 10.00 & 10.00 & 10.00 & 10.00 & 9.94 & 10.00 & 10.00 & 10.00 & 10.00 \\
 & \cTwo & \cellcolor{gray!15} & 10.00 & 9.94 & 10.00 & 10.00 & 9.94 & 10.00 & 10.00 & 10.00 & 10.00 \\
 & \cThree & \cellcolor{gray!15} & 10.00 & 9.88 & 10.00 & 10.00 & 9.96 & 10.00 & 10.00 & \textbf{9.76**} & 10.00 \\
 & \comp & \cellcolor{gray!15} & 10.00 & \textbf{7.56***} & \textbf{4.92***} & \textbf{9.54***} & \textbf{8.28***} & \textbf{8.24***} & \textbf{8.66***} & \textbf{8.74***} & \textbf{8.18***} \\
\midrule
\multirow{4}{*}{Narrative Free Recall}
 & \cOne & 0.604 & 0.997 & 0.756 & 0.683 & 0.935 & 1.000 & 1.000 & 0.999 & 1.000 & 1.000 \\
 & \cTwo & \cellcolor{gray!15} & 0.9967 & \textbf{0.647**} & 0.658 & \textbf{0.871**} & \textbf{0.808**} & 0.695 & \textbf{0.996**} & 0.9999 & \textbf{0.772**} \\
 & \cThree & \cellcolor{gray!15} & \textbf{0.9931**} & \textbf{0.675**} & 0.657 & \textbf{0.777**} & \textbf{0.684**} & \textbf{0.604*} & \textbf{0.983**} & \textbf{0.809**} & \textbf{0.699**} \\
 & \comp & \cellcolor{gray!15} & \textbf{0.7364***} & \textbf{0.6015***} & \textbf{0.5026***} & \textbf{0.6559***} & \textbf{0.5720***} & \textbf{0.5657***} & \textbf{0.5817***} & \textbf{0.5063***} & \textbf{0.5680***} \\
\bottomrule
 
\end{tabular}
}
\end{table*}

\begin{table*}[t]
\centering
\small
\caption{Human--model similarity by task, model, and prompt condition (measured as $1-$ normalized Wasserstein distance). \cOne is the baseline. $\Delta$\cTwo, $\Delta$\cThree, and $\Delta$\comp denote changes relative to \cOne. Statistically significant differences are shown in bold.}
\label{tab:wasserstein-norm-delta}
\setlength{\tabcolsep}{3pt}
\vspace{0.1cm}
\resizebox{\textwidth}{!}{%
\begin{tabular}{llrrrrrrrrr}
\toprule
Task & Cond. & Claude Opus 4.6 & Llama 3 8B & Llama 3.3 70B & GPT-5.4 & Qwen3 8B NT & Qwen3 8B T & Qwen3 30B Inst & Qwen3 30B Think & Qwen3 Next 80B \\
\midrule
\multirow{4}{*}{Digit Span} 
& \cOne & 0.324 & 0.324 & 0.324 & 0.324 & 0.324 & 0.324 & 0.324 & 0.324 & 0.324 \\
& $\Delta$\cTwo & 0.000 & 0.000 & 0.000 & 0.000 & 0.000 & 0.000 & 0.000 & +0.005 & 0.000 \\
& $\Delta$\cThree & 0.000 & 0.000 & 0.000 & 0.000 & \textbf{+0.353**} & +0.387 & 0.000 & \textbf{+0.550**} & +0.030 \\
& $\Delta$\comp & 0.000 & \textbf{+0.353*} & \textbf{+0.607*} & \textbf{+0.050*} & \textbf{+0.586*} & \textbf{+0.607*} & \textbf{+0.050*} & \textbf{+0.551*} & \textbf{+0.551*} \\
\midrule

\multirow{4}{*}{Reverse Digit Span} 
& \cOne & 0.276 & 0.843 & 0.309 & 0.276 & 0.276 & 0.276 & 0.305 & 0.276 & 0.282 \\
& $\Delta$\cTwo & 0.000 & \textbf{+0.096**} & +0.026 & 0.000 & 0.000 & 0.000 & +0.018 & 0.000 & +0.013 \\
& $\Delta$\cThree & 0.000 & \textbf{+0.077**} & +0.013 & +0.008 & \textbf{+0.059**} & 0.000 & +0.024 & \textbf{+0.069*} & +0.003 \\
& $\Delta$\comp & \textbf{+0.150*} & \textbf{-0.118*} & \textbf{+0.416*} & \textbf{+0.400*} & \textbf{+0.572*} & \textbf{+0.616*} & \textbf{+0.421*} & \textbf{+0.616*} & \textbf{+0.662*} \\
\midrule

\multirow{4}{*}{N-Back} 
& \cOne & 0.887 & 0.855 & 0.451 & 0.925 & 0.863 & 0.858 & 0.142 & 0.859 & 0.959 \\
& $\Delta$\cTwo & \textbf{+0.015*} & \textbf{-0.072**} & \textbf{+0.301**} & +0.009 & +0.005 & +0.006 & 0.000 & -0.001 & 0.000 \\
& $\Delta$\cThree & \textbf{+0.016*} & \textbf{+0.077**} & \textbf{+0.379**} & +0.010 & -0.003 & +0.006 & 0.000 & -0.001 & -0.002 \\
& $\Delta$\comp & \textbf{-0.013*} & +0.028 & \textbf{+0.524***} & \textbf{-0.014**} & \textbf{+0.105***} & \textbf{+0.112***} & \textbf{+0.832***} & \textbf{+0.078***} & \textbf{+0.019**} \\
\midrule

\multirow{4}{*}{Word Recognition} 
& \cOne & 0.333 & 0.891 & 0.333 & 0.333 & 0.333 & 0.333 & 0.333 & 0.333 & 0.333 \\
& $\Delta$\cTwo & 0.000 & +0.027 & 0.000 & 0.000 & 0.000 & 0.000 & 0.000 & +0.006 & 0.000 \\
& $\Delta$\cThree & 0.000 & +0.020 & 0.000 & 0.000 & +0.015 & 0.000 & +0.004 & \textbf{+0.450**} & 0.000 \\
& $\Delta$\comp & \textbf{+0.217***} & \textbf{-0.023***} & \textbf{+0.236***} & \textbf{+0.227***} & \textbf{+0.255***} & \textbf{+0.248***} & \textbf{+0.063**} & \textbf{+0.210***} & \textbf{+0.230***} \\
\midrule

\multirow{4}{*}{Variable Mapping} 
& \cOne & 0.635 & 0.635 & 0.635 & 0.765 & 0.635 & 0.635 & 0.635 & 0.796 & 0.912 \\
& $\Delta$\cTwo & 0.000 & 0.000 & 0.000 & \textbf{-0.126**} & 0.000 & 0.000 & 0.000 & \textbf{+0.075**} & \textbf{-0.277**} \\
& $\Delta$\cThree & 0.000 & 0.000 & 0.000 & \textbf{-0.131**} & 0.000 & 0.000 & 0.000 & \textbf{+0.060**} & \textbf{-0.264**} \\
& $\Delta$\comp & 0.000 & +0.015 & 0.000 & +0.019 & \textbf{+0.152***} & \textbf{+0.194***} & 0.000 & \textbf{-0.161***} & \textbf{-0.277***} \\
\midrule

\multirow{4}{*}{Map Task} 
& \cOne & 0.702 & 0.860 & 0.816 & 0.702 & 0.702 & 0.702 & 0.702 & 0.702 & 0.702 \\
& $\Delta$\cTwo & 0.000 & +0.012 & \textbf{-0.060**} & 0.000 & 0.000 & 0.000 & +0.002 & +0.007 & 0.000 \\
& $\Delta$\cThree & 0.000 & \textbf{+0.032**} & -0.004 & 0.000 & 0.000 & 0.000 & +0.008 & \textbf{+0.228**} & 0.000 \\
& $\Delta$\comp & \textbf{+0.042*} & \textbf{-0.048***} & \textbf{+0.157***} & +0.008 & \textbf{+0.102***} & \textbf{+0.103***} & \textbf{+0.219***} & \textbf{+0.087***} & \textbf{+0.027**} \\
\midrule

\multirow{4}{*}{Craft Task} 
& \cOne & 0.811 & 0.862 & 0.813 & 0.811 & 0.821 & 0.831 & 0.811 & 0.837 & 0.811 \\
& $\Delta$\cTwo & 0.000 & \textbf{-0.104**} & -0.002 & \textbf{+0.016*} & 0.000 & -0.020 & 0.000 & -0.001 & 0.000 \\
& $\Delta$\cThree & 0.000 & \textbf{-0.140**} & +0.002 & \textbf{+0.085**} & -0.010 & -0.020 & 0.000 & \textbf{+0.103**} & 0.000 \\
& $\Delta$\comp & 0.000 & \textbf{-0.066**} & \textbf{+0.099***} & +0.004 & \textbf{+0.071***} & \textbf{+0.052***} & \textbf{+0.054***} & \textbf{+0.053***} & \textbf{+0.081***} \\
\midrule

\multirow{4}{*}{Narrative QA} 
& \cOne & 0.793 & 0.873 & 0.796 & 0.793 & 0.796 & 0.793 & 0.798 & 0.801 & 0.793 \\
& $\Delta$\cTwo & +0.021 & +0.034 & \textbf{+0.039*} & +0.021 & +0.024 & \textbf{+0.060**} & +0.020 & +0.039 & +0.021 \\
& $\Delta$\cThree & -0.015 & +0.032 & -0.003 & -0.015 & +0.003 & +0.030 & -0.012 & +0.060 & -0.015 \\
& $\Delta$\comp & \textbf{+0.037*} & \textbf{-0.008***} & \textbf{-0.027***} & \textbf{+0.129***} & \textbf{+0.130***} & \textbf{+0.143***} & \textbf{+0.157***} & \textbf{+0.117***} & \textbf{+0.141***} \\
\midrule

\multirow{4}{*}{Factual QA} 
& \cOne & 0.663 & 0.663 & 0.663 & 0.663 & 0.672 & 0.663 & 0.663 & 0.663 & 0.663 \\
& $\Delta$\cTwo & 0.000 & +0.009 & 0.000 & 0.000 & 0.000 & 0.000 & 0.000 & 0.000 & 0.000 \\
& $\Delta$\cThree & 0.000 & +0.018 & 0.000 & 0.000 & -0.003 & 0.000 & 0.000 & \textbf{+0.034**} & 0.000 \\
& $\Delta$\comp & 0.000 & \textbf{+0.287***} & \textbf{+0.106***} & \textbf{+0.064***} & \textbf{+0.204***} & \textbf{+0.212***} & \textbf{+0.167***} & \textbf{+0.140***} & \textbf{+0.211***} \\
\midrule

\multirow{4}{*}{Narrative Free Recall} 
& \cOne & 0.607 & 0.849 & 0.921 & 0.669 & 0.604 & 0.604 & 0.605 & 0.604 & 0.604 \\
& $\Delta$\cTwo & 0.000 & \textbf{+0.102**} & +0.008 & \textbf{+0.064**} & \textbf{+0.193**} & +0.304 & \textbf{+0.003**} & 0.000 & \textbf{+0.228**} \\
& $\Delta$\cThree & \textbf{+0.002**} & \textbf{+0.079**} & +0.023 & \textbf{+0.159**} & \textbf{+0.316**} & \textbf{+0.332*} & \textbf{+0.016**} & \textbf{+0.191**} & \textbf{+0.302**} \\
& $\Delta$\comp & \textbf{+0.261***} & \textbf{+0.118***} & \textbf{-0.022***} & \textbf{+0.279***} & \textbf{+0.345***} & \textbf{+0.339***} & \textbf{+0.353***} & \textbf{+0.295***} & \textbf{+0.347***} \\

\midrule
\multirow{4}{*}{\textbf{Average}}
& \cOne 
& 0.603 & 0.766 & 0.606 & 0.626 & 0.603 & 0.602 & 0.602 & 0.620 & 0.623 \\
& $\Delta$\cTwo
& +0.004 & +0.010 & +0.031 & -0.002 & +0.022 & +0.035 & +0.004 & +0.013 & -0.002 \\
& $\Delta$\cThree
& +0.000 & +0.016 & +0.043 & +0.012 & +0.073 & +0.074 & +0.004 & +0.133 & +0.006 \\
& $\Delta$\comp
& +0.069 & +0.054 & +0.209 & +0.117 & +0.252 & +0.263 & +0.232 & +0.199 & +0.199 \\
\bottomrule
\end{tabular}%
}
\end{table*}

\end{document}